\pdfoutput=1
\documentclass[11pt]{article}

\usepackage{EMNLP2023}

\usepackage{times}
\usepackage{latexsym}

\usepackage{amssymb}% http://ctan.org/pkg/amssymb
\usepackage{pifont}% http://ctan.org/pkg/pifont

\usepackage[T1]{fontenc}
\usepackage[utf8]{inputenc}

\usepackage{microtype}

\usepackage{booktabs} 
\usepackage{multirow}
\usepackage{graphicx}
\usepackage{arydshln}
\usepackage{amsfonts}
\usepackage{amsmath}
\usepackage{amssymb}
\usepackage{inconsolata}

\title{Masked Path Modeling for Vision-and-Language Navigation}

\author{Zi-Yi Dou$^{\dagger}$, Feng Gao$^{\sharp}$, Nanyun Peng$^{\dagger}$\\
   $^{\dagger}$University of California, Los Angeles \; $^{\sharp}$Amazon Alexa AI \\
  {\tt \{zdou,violetpeng\}@cs.ucla.edu }   fenggo@amazon.com }

\begin{document}
\maketitle
\begin{abstract}
Vision-and-language navigation (VLN) agents are trained to navigate in real-world environments by following natural language instructions. A major challenge in VLN is the limited availability of training data, which hinders the models' ability to generalize effectively. Previous approaches have attempted to address this issue by introducing additional supervision during training, often requiring costly human-annotated data that restricts scalability. %Although several pretraining objectives have been proposed for VLN,\vp{motivate your approach by ``prior works tried pretrained but their pretraining objectives are not as good'' seems to be a weak/indirect motivation. Should probably say what hinders the current VLN models from performing well. E.g., small data size makes it hard to generalize, and pretraining usually requires human annotation(?)} 
 %but they often have limitations such as relying on a limited number of human-annotated data, which hampers scalability, or focusing on representation learning while not explicitly on improving the model's ability to generate actions. 
In this paper, we introduce a masked path modeling (MPM) objective, which pretrains an agent using self-collected data for downstream navigation tasks. Our proposed method involves allowing the agent to actively explore navigation environments without a specific goal and collect the paths it traverses. Subsequently, we train the agent on this collected data to reconstruct the original path given a randomly masked subpath. This way, the agent can actively accumulate a diverse and substantial amount of data while learning conditional action generation. %\vp{it's unclear how the statement after ``this way'' is achieved. how did you connect MPM to instructions?} 
To evaluate the effectiveness of our technique, we conduct experiments on various VLN datasets and demonstrate the versatility of MPM across different levels of instruction complexity. Our results exhibit significant improvements in success rates, with enhancements of 1.32\%, 1.05\%, and 1.19\% on the val-unseen split of the Room-to-Room, Room-for-Room, and Room-across-Room datasets, respectively. Furthermore, we conduct an analysis that highlights the potential for additional improvements when the agent is allowed to explore unseen environments prior to testing. 
 
\end{abstract}

\section{Introduction}

A vision-and-language navigation (VLN) agent is trained to interpret natural language instructions and navigate within an environment to achieve a specified goal. This task requires the agent to possess several sophisticated abilities, including understanding and grounding language phrases to visual objects, as well as planning and executing actions in a real-world setting. To address these challenges, researchers have adopted the pretraining-then-finetuning paradigm, which has been successful in natural language processing~\cite{elmo,kenton2019bert} as well as in computer vision~\cite{chen2021empirical,he2022masked}. By utilizing various supervision signals and proposing pretraining objectives, significant improvements have been demonstrated across VLN tasks.

\begin{figure}[t]
 \centering
 \includegraphics[width=0.49\textwidth]{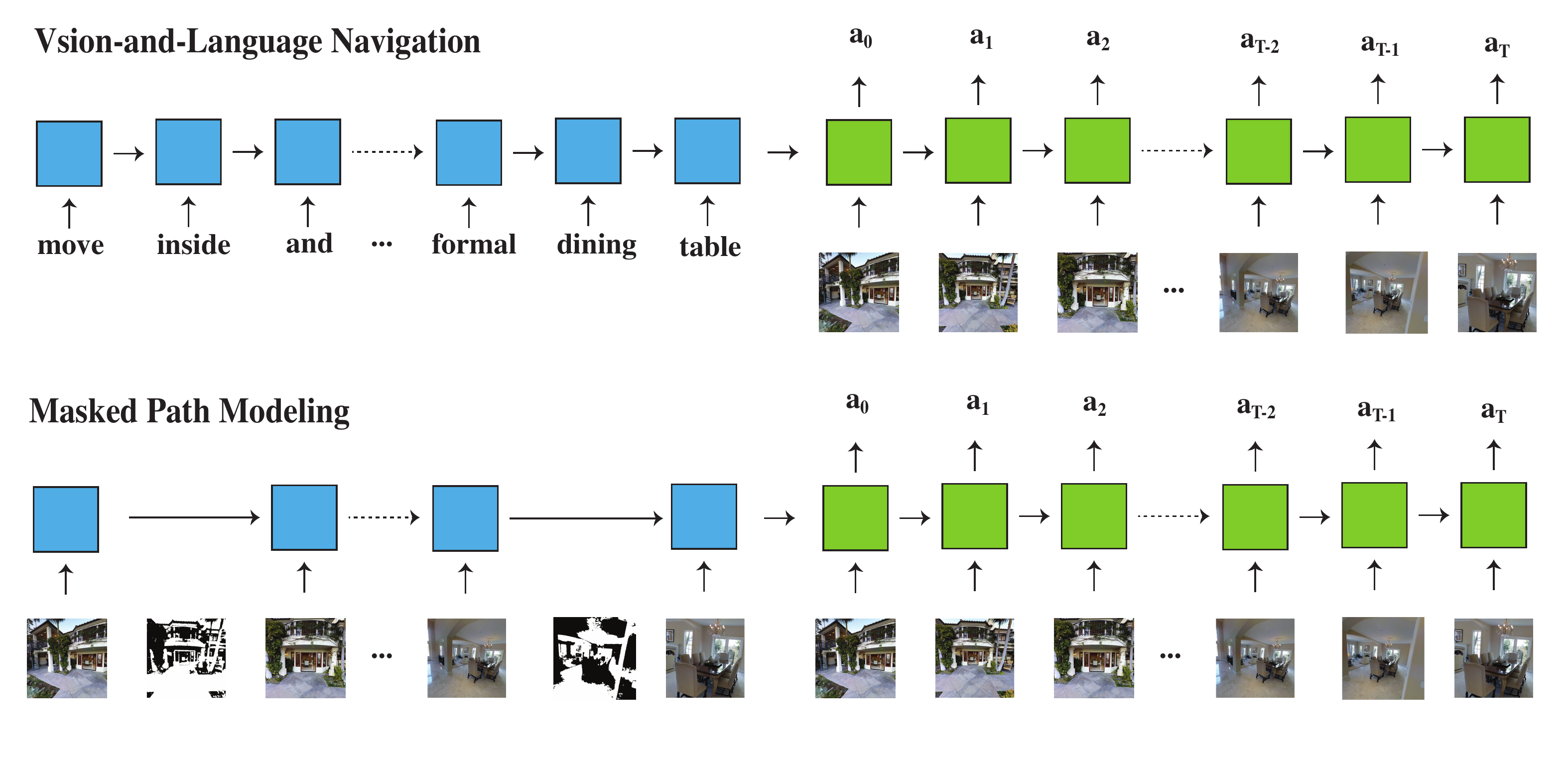}
 \caption{Vision-and-language navigation agents are trained to predict actions in a real-world environment given a natural language instruction. We incorporate our proposed masked path modeling objective into VLN training, where the agent is trained to reconstruct the complete path given a randomly masked subpath. The figure is adapted from~\citet{anderson2018vision}.}
 \label{fig:intro}
\end{figure}

Pretraining models on large web data has emerged as an effective approach for representation learning. In the field of VLN, researchers have explored the use of internet-scale image-text datasets to acquire grounded vision and language representations. Notable contributions in this area include the works of ~\citet{majumdar2020improving} and~\citet{guhur2021airbert}, who leverage web-scraped image-caption corpora to learn general vision-language representations and then finetune the models using in-domain VLN data to adapt the representations specifically for VLN tasks. Similarly,~\citet{shen2022much} and~\citet{khandelwal2022simple} employ the CLIP vision encoder~\cite{radford2021learning} pretrained with image-text contrastive loss and showcase the application of such models in various embodied tasks. While the utilization of vision-language pretrained representations improves the alignment between vision and language modalities, it is important to note that these models do not explicitly optimize for \textit{generating actions} in real-world environments, i.e. the models are unaware of how to connect the learned \textit{representations to actions}, which is a critical skill for VLN tasks.

In order to integrate action generation goals into the pretraining process, PREVALENT~\cite{hao2020towards} employs a combination of human-annotated and synthetic image-text-action triplets. This approach incorporates two main objectives: a single-step action prediction objective and a masked instruction modeling objective. Subsequent research, such as the work conducted by~\citet{chen2021history}, builds upon and improves these objectives. While their agents are trained to generate actions during pretraining, the scalability of the pretraining objective is limited due to the scarcity of annotated instruction-action pairs. Specifically, they require training on pairs of natural language instruction along with action sequences which are costly to obtain in a scalable way.

%In this paper, we present an approach \vp{to pretrain VLN models using MPM, which leverages xxx} to address the limitations of previous work in VLN. Our proposed objective, called masked path modeling (MPM), 
In this paper, we present an approach to pretrain VLN models with masked path modeling (MPM), which leverages in-domain data actively collected by an agent for self-supervised learning. The proposed objective targets addressing the two major limitations of previous work:
\begin{itemize}
    \item To collect scalable pretraining data, during the pretraining phase, the VLN agent explores the environment randomly and gathers navigation paths, which are then used for pretraining the agent. Because the agent actively explores different environments, we can actively collect a large number of diverse paths.
    \item To optimize the utilization of MPM for the VLN agent, MPM explicitly focuses on conditional action generation. Concretely, as shown in Figure~\ref{fig:intro}, to construct the MPM objective, we randomly mask certain viewpoints in the self-collected paths and train the agent to reconstruct the original paths based on the masked ones. MPM is similar to the VLN objective, with the distinction that the instructions are presented as masked paths rather than natural language instructions. Consequently, MPM effectively prepares the agent for tasks requiring conditional action generation. 
    %\item To optimize the utilization of MPM for the VLN agent, MPM explicitly focuses on training the agent to effectively follow embodied instructions and generate actions accordingly. Concretely, as shown in Figure~\ref{fig:intro}, to construct the MPM objective, we randomly mask certain viewpoints in the self-collected paths and train the agent to reconstruct the original paths based on the masked ones. MPM closely resembles the VLN objective, with the distinction that the instructions are presented as masked paths rather than natural language instructions. Consequently, MPM effectively prepares the agent for tasks requiring conditioned action generation. 
    %\vp{I don't think this is super convincing, but I haven't gone over your experiments yet, maybe you have experiment to verify this? i.e. masked path resembles instructions. E.g., did you share the parameters for the text and path encoders? or did you explicitly try to align the latent space?} \zd{i've changed the wording. yes most of the model parameters are shared (as in figure 2). }
\end{itemize} 
As a result, our pretraining objective is scalable and well-suited for addressing the VLN task.

We evaluate the proposed method on various VLN datasets with different types of instructions, including Room-to-Room~\cite{r2r}, Room-for-Room~\cite{jain2019stay}, and Room-across-Room~\cite{ku2020room}. Experimental results demonstrate that MPM can achieve significant improvements in both seen and unseen environments compared with strong baselines. For example, we achieve improvements of 1.32\%, 1.05\%, and 1.19\% on the val-unseen split of the Room-to-Room, Room-for-Room, and Room-across-Room datasets, respectively. Furthermore, an analysis reveals the potential for additional improvements when the agent is allowed to explore unseen environments prior to testing.

\begin{figure*}[ht]
 \centering
 \includegraphics[width=1.0\textwidth]{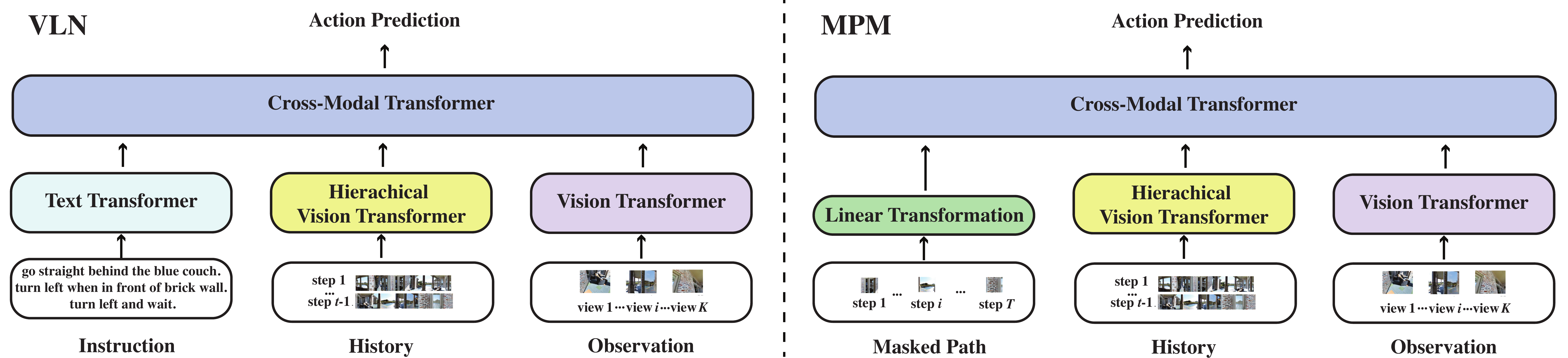}
 \caption{We follow~\citet{chen2021history} to design the base model architecture. In VLN, the model separately encodes the given instruction with a text encoder, its past history with a hierarchical visual encoder, and its current observations with another vision encoder; the encoded representations are then fed into a joint cross-modal transformer to predict the final action. In MPM, we directly feed a masked subpath to the cross-modal transformer instead of a language instruction and the model is trained to predict the original action given the masked subpath. All the parameters between VLN and MPM are shared.}
 \label{fig:arch}
\end{figure*}

\section{Methods}
In this section, we first introduce the basic settings and model architecture of vision-and-language navigation, then illustrate the details of each component of our proposed approach.

\subsection{Background}

\paragraph{Training Data.} The training data $\mathcal{D}$ of VLN consists of parallel instruction-action pairs $\{(\mathbf{i}^k, \mathbf{a}^k)\}$ from different environments. However, it is hard to manually annotate a large amount of instruction-action data for VLN. Therefore, researchers have proposed various data augmentation strategies~\cite{fried2018speaker,tan2019learning,li2022envedit,kamath2023new} to provide additional supervision for VLN models. For example, a common practice is to train a speaker model~\cite{fried2018speaker} to generate language instructions given randomly sampled paths and the generated instruction-action pairs can be used to enrich the training data for VLN training. 

\paragraph{Base Settings.} Using the parallel instruction-action pairs $\mathcal{D}$, a VLN agent is trained to follow a given language instruction and generate a sequence of actions to reach the final destination in a photo-realistic environment. Formally, in a given environment $\mathbf{e}$, the navigation agent parameterized by $\theta$ learns to model the distribution $P(\mathbf{a}|\mathbf{i}; \theta)$, where $\mathbf{i}$ and $\mathbf{a}$ denote instruction and action variables, respectively.

\paragraph{Model Architecture.} In this paper, we employ a history-ware multimodal transformer architecture design following~\citet{chen2021history}, although it should be noted that our approach is in principle compatible with any existing sequence-to-sequence architectures in VLN. Specifically, as shown in Figure~\ref{fig:arch}, at each action prediction step, we have a transformer text encoder to encode the given language instruction, a hierarchical vision transformer to encode all the past observations of the agent, and another vision transformer to encoder the agent panoramic observation of the current step; then, the three types of representations will be concatenated together and fed into a cross-modal transformer for joint encoding, and the final pooled representation is used for action prediction. 

\paragraph{Text Features.} Following previous work~\cite{chen2021history}, we use pretrained text encoder to encode the language instructions, and its weights are finetuned during VLN training. We use the standard BERT model~\cite{kenton2019bert} for the R2R and R4R datasets~\cite{r2r,jain2019stay} and the XLM-R model~\cite{conneau2020unsupervised} to encode multilingual instructions for the RxR dataset~\cite{ku2020room} because R2R and R4R consist of only English instructions whereas RxR contains multilingual instructions.

\paragraph{Vision Features.} At each step, the agent is given a panoramic observation of its current position in the environment, denoted as $\{v_i\}_{i=1}^K$. For each view in the panoramic observation, its vision feature is first extracted using a pretrained vision encoder. While many of the previous methods~\cite{anderson2018vision,chen2021history} use vision encoders pretrained on ImageNet~\cite{fei2009imagenet} for image classification, we find that CLIP vision encoder~\cite{radford2021learning} achieves stronger performance as in ~\citet{shen2022much} and we choose to use CLIP to first extract vision features and then the CLIP features are fed into the transformers for history and observation encoding. 

For the current observations, in addition to the CLIP features, we also feed the model with the relative angle of each view $v_i$ in the panoramic observation, represented as $(sin \theta_i, cos \theta_i, sin \phi_i, cos \phi_i)$  where $\theta_i$ and $\phi_i$ are the relative heading
and elevation angle to the agent’s orientation, respectively. The panoramic representations $\{\text{CLIP}(v_i)\}_{i=1}^K$ are then fed into a transformer model to obtain $K$ processed representations.

The model also keeps track of its past observations with a hierarchical vision transformer, where the panoramic observation at each step is first encoded by a single vector with a vision transformer, and all the panoramic representations are jointly encoded with another transformer along the temporal dimension.

\paragraph{Cross-Modal Interactions.} The history features and the current observation features are concatenated as the vision modality, and a dual-stream cross-modal fusion architecture is used to encode both the vision and text modalities and allow for cross-modal information exchange. At each layer, we have a self-attention block for inter-modal interactions and a cross-attention block for vision-text interactions.

\subsection{Masked Path Modeling}
In this part, we illustrate the main idea of our proposed masked path modeling method. We go through the details of the active data collection, model architecture, and training strategies.

\paragraph{General Framework.} Masked path modeling is inspired by the masked data modeling pretraining methods in the language and vision communities~\cite{kenton2019bert,he2022masked}, where the general idea is that the model is trained to reconstruct an original input (e.g., a sentence or an image) given parts of the input masked. In VLN, we follow this paradigm and propose to first ask an agent to perform a sequence of actions and collect a path consisting of several viewpoints $\mathbf{p}=\langle p_1, p_2, \cdots, p_n \rangle$, then we mask 25\% of the viewpoints in this path and feed the observations along the masked path $\mathbf{p}_{m} = \langle p_{m_1}, p_{m_2}, \cdots, p_{m_k} \rangle$ to the agent and the agent is trained to perform the same sequence of actions as before to reconstruct the original path $\mathbf{p}$ given $\mathbf{p}_{m}$.

\paragraph{Data Collection.} One of the major bottlenecks of VLN is the lack of training data and it is hard to collect large-scale human-annotated data for VLN. In masked path modeling, however, the agent can actively collect a great amount of data given an environment for training. During the data collection period, we ask the agent to randomly choose the next viewpoint with equal probabilities at each step. Also, we keep track of all the visited viewpoints, and the agent is not allowed to visit the same viewpoint twice. To control the length of the paths, we pre-compute the statistics of agent paths in the training data and make the lengths of the sampled paths follow the distribution of that in the training data. More sophisticated path collection techniques such as using techniques to encourage the diversity of sampled paths may also be used but here we leave it as future work.

\paragraph{Model Architecture.} As in Figure~\ref{fig:arch}, the main difference between masked path modeling and the VLN objective is that the input of masked path modeling is a sequence of visual observations instead of a natural language instruction. Therefore, we employ the same architecture as the original HAMT model except that we perform a linear transformation on the CLIP-encoded visual features so as to match the input and model dimensions, and then directly feed transformed features to the crossmodal transformer module. While collecting the visual features along a masked path, we do not use the panoramic view but only the view that the agent currently faces so as to make the pretraining task harder.\footnote{During decoding, the agent still receives panoramic views as in previous work~\cite{fried2018speaker,chen2021history}.} All the module parameters are shared between the masked path modeling and VLN objectives.

\paragraph{Training Strategies.} We include our masked path modeling objective into the pretraining and finetuning stages of HAMT~\cite{chen2021history}. Concretely, during pretraining, the agent is jointly pretrained with masked path modeling and standard objectives including masked language modeling and instruction trajectory matching. We also include single-step action prediction and regression (SAP/SAR), and spatial relationship prediction (SPREL) objectives as in~\citet{chen2021history}.\footnote{We do not use the masked region modeling objective in HAMT because it requires distilling the knowledge of an image classification model while our vision encoder is CLIP. Also, the CLIP vision encoder is frozen during pretraining to save computation time.} The SAP and SAR objectives ask the model to predict the next action based on instruction, history from the ground-truth demonstration, and the current observation with imitation learning, where SAP formulates the task as a classification task while SAR trains the model to regress the action heading and elevation angles. The SPREL objective trains the model to predict the relative spatial position of two views in a panorama based
on only visual features, angle features, or both types of features. We refer readers to~\citet{chen2021history} for more details. %\vp{this assumes your readers are familiar with HAMT objectives, but you didn't provide a recap in your background and the expectation is probably too high} \zd{fixed} 

Then, during finetuning, the model is jointly trained with both masked path modeling and the VLN objective with equal loss weights.\footnote{We did not see significant performance differences when tuning the loss weights in our preliminary studies.} %\vp{this seems to be a ``magic number'', have you tried different weights? why not?} 
We combine the Asynchronous Advantage Actor-Critic (A3C) reinforcement learning objective~\cite{mnih2016asynchronous} and imitation learning objective for the VLN objective following previous work~\cite{tan2019learning,chen2021empirical}, but only use the imitation learning objective for masked path modeling because it is stable and also it is non-trivial to design step-wise rewards in this setting.

\begin{table*}[t]
\small
 \renewcommand{\tabcolsep}{1.7mm}
\centering
  \begin{tabular}{@{}lccccccccccccccc@{}}
    \toprule
     \multirow{2}*{\bf Model}      & \multicolumn{4}{c@{}}{\bf Validation Seen}  & \multicolumn{4}{c@{}}{\bf Validation Unseen} & \multicolumn{4}{c@{}}{\bf Test Unseen}                                   \\ 
           & \bf TL & \bf  NE$\downarrow$ & \bf  SR$\uparrow$ & \bf SPL$\uparrow$ & \bf TL & \bf  NE$\downarrow$ & \bf  SR$\uparrow$ & \bf SPL$\uparrow$ & \bf TL & \bf  NE$\downarrow$ & \bf  SR$\uparrow$ & \bf SPL$\uparrow$  \\
           \midrule
HAMT~\cite{chen2021history} & 11.15 & 2.51 & 76 & 72 & 11.46 & \bf 2.29 & 66  & 61 & 12.27 & 3.93 & 65 & 60 \\
HAMT+ & 11.11 & 2.65 & 75.02 & 71.75 & 11.93 & 3.34 & 67.05 & 61.69 & 12.70 & 3.57 & 67.19 & 61.94\\
HAMT+ w/ MPM & 10.86 & \bf 2.43 & \bf 76.30 & \bf 72.85  & 11.99 & 3.44 & \bf 68.37 & \bf 62.59 & 12.54 & \bf 3.47 & \bf 67.79 & \bf 62.54 \\
\bottomrule
\end{tabular}
\caption{Results on the Room-to-Room dataset~\cite{r2r}. We incorporate MPM into a strong baseline (HAMT+) and achieve significant improvements across settings. The best scores are in \textbf{bold}.}
\label{tab:r2r}
\end{table*}

\begin{table*}[t]
\small
 \renewcommand{\tabcolsep}{1.8mm}
\centering
  \begin{tabular}{@{}lccccccccccccccc@{}}
    \toprule
     \multirow{2}*{\bf Model}      & \multicolumn{5}{c@{}}{\bf Validation Seen}  & \multicolumn{5}{c@{}}{\bf Validation Unseen}                                   \\ 
           %& \bf TL & \bf  NE$\downarrow$ & \bf  SR$\uparrow$ & \bf SPL$\uparrow$ & \bf CLS$\uparrow$ & \bf nDTW$\uparrow$ & \bf sDTW$\uparrow$ &  \bf TL & \bf  NE$\downarrow$ & \bf  SR$\uparrow$ & \bf SPL$\uparrow$ & \bf CLS$\uparrow$ & \bf nDTW$\uparrow$ & \bf sDTW$\uparrow$  \\
           & \bf  NE$\downarrow$ & \bf  SR$\uparrow$  & \bf CLS$\uparrow$ & \bf nDTW$\uparrow$ & \bf sDTW$\uparrow$ &  \bf  NE$\downarrow$ & \bf  SR$\uparrow$ & \bf CLS$\uparrow$ & \bf nDTW$\uparrow$ & \bf sDTW$\uparrow$  \\
           \midrule
HAMT~\cite{chen2021history} & - & - & -  & - & -   & 6.09 & 44.6  & 57.7 & 50.3 & 31.8 \\
HAMT+  & 4.62 & 57.29 & 67.97 & 61.01 & 41.96  & 5.90 & 44.75  & 61.84 & 54.18 & 33.89 \\
HAMT+ w/ MPM  & \bf 4.29 & \bf 59.13  & \bf 70.50 & \bf 64.88 & \bf 48.28    & \bf 5.65 & \bf 46.88   & \bf 62.76 & \bf 55.23 & \bf 35.50\\
\bottomrule
\end{tabular}
\caption{Results on the Room-for-Room dataset~\cite{jain2019stay}. MPM can also improve the model performance in this setting across all the evaluation metrics. The best scores are in \textbf{bold}.}
\label{tab:r4r}
\end{table*}

\begin{table*}[t]
\small
 \renewcommand{\tabcolsep}{0.8mm}
\centering
  \begin{tabular}{@{}lccccccccccccccc@{}}
    \toprule
     \multirow{2}*{\bf Model}      & \multicolumn{4}{c@{}}{\bf Validation Seen}  & \multicolumn{4}{c@{}}{\bf Validation Unseen} & \multicolumn{4}{c@{}}{\bf Test Unseen}                                   \\ 
           & SR$\uparrow$ &  SPL$\uparrow$ & nDTW$\uparrow$ & sDTW$\uparrow$  & SR$\uparrow$ &  SPL$\uparrow$ & nDTW$\uparrow$ & sDTW$\uparrow$  & SR$\uparrow$ &  SPL$\uparrow$ & nDTW$\uparrow$ & sDTW$\uparrow$  \\
           \midrule
HAMT~\cite{chen2021history}  & 59.4 & 58.9 & 65.3 & 50.9 & 56.5 & 56.0&  63.1 & 48.3 & 53.12 & 46.62 & 59.94 & 45.19 \\
HAMT+ & 63.93 & 59.93 & 68.59 & 55.47 & 62.00 & 58.05 & 67.52 & 53.87 & - & - & - & - \\
HAMT+ w/ MPM  &  \bf 67.73 & \bf 63.89 & \bf 71.02 & \bf 58.86 & \bf 63.51 & \bf 59.24 & \bf 67.71 & \bf 54.53 &  \bf 60.00 & \bf 52.52 & \bf 63.97 & \bf 51.13\\
\bottomrule
\end{tabular}
\caption{Results on the Room-across-Room dataset~\cite{ku2020room}. The best scores are in \textbf{bold}.}
\label{tab:rxr}
\end{table*}

\begin{table*}[t]
\small
\centering
  \begin{tabular}{@{}lccccccccccccccc@{}}
    \toprule
     \multirow{2}*{\bf Model}      & \multicolumn{4}{c@{}}{\bf R2R Validation Unseen}            & \multicolumn{5}{c@{}}{\bf R4R Validation Unseen}                             \\ 
           & \bf TL & \bf  NE$\downarrow$ & \bf  SR$\uparrow$ & \bf SPL$\uparrow$  &  \bf  NE$\downarrow$ & \bf  SR$\uparrow$ & \bf CLS$\uparrow$ & \bf nDTW$\uparrow$ & \bf sDTW$\uparrow$  \\
           \midrule
%HAMT+  & 11.93 & 3.34 & 67.05 & 61.69 & 5.90 & 44.75  & 61.84 & 54.18 & 33.89  \\
HAMT+ w/ MPM  & 11.99 & 3.44 & 68.37 & 62.59 & 5.65 & 46.88   & 62.76 & 55.23 & 35.50 \\
HAMT+ w/ MPM-Prexplore & 11.37 & \bf 3.33 & \bf 69.60 & \bf 64.69 &  \bf 5.13 & \bf 51.34 & \bf 63.72 & \bf 57.95 & \bf 39.06 \\
\bottomrule
\end{tabular}
\caption{Pre-exploring the test environments with MPM can further improve the model performance.}
\label{tab:explore}
\end{table*}

\iffalse
\begin{table*}[t]
\small
\centering
  \begin{tabular}{@{}lccccccccccccccc@{}}
    \toprule
     \multirow{2}*{\bf Mask Ratio (\%)}      & \multicolumn{4}{c@{}}{\bf Validation Seen}  & \multicolumn{4}{c@{}}{\bf Validation Unseen}                                   \\ 
           & \bf TL & \bf  NE$\downarrow$ & \bf  SR$\uparrow$ & \bf SPL$\uparrow$ &  \bf TL & \bf  NE$\downarrow$ & \bf  SR$\uparrow$ & \bf SPL$\uparrow$  \\
           \midrule
           0  &  10.98 & 2.38 & 76.89 & 73.63 & 11.40 & 3.40& 67.43 & 61.90 \\
 25 & 10.86 & 2.43 & 76.30 & 72.85  & 11.99 & 3.44 & 68.37 & 62.59 \\
 50 & 11.03 & 2.61 & 74.73 & 71.63 & 11.59 & 3.35 & 67.18 & 61.97 \\
 75 & 11.28 & 2.49 & 75.32 & 71.44 & 12.15 & 3.42 & 66.58 & 60.42\\
100  &  11.19 & 2.48 & 76.00 & 72.43& 11.94 & 3.44 & 67.90 & 63.22\\
\bottomrule
\end{tabular}
\caption{Effect of the mask ratio of the MPM objective. The model is generally robust to this hyper-parameter, with 25\% achieving the best performance on unseen environments. }
\label{tab:mask}
\end{table*}
\fi

\begin{table*}[t]
\small
\centering
  \begin{tabular}{@{}lccccccccccccccc@{}}
    \toprule
     \multirow{2}*{\bf Path Design}      & \multicolumn{4}{c@{}}{\bf R2R Validation Unseen}            & \multicolumn{5}{c@{}}{\bf R4R Validation Unseen}                             \\ 
           & \bf TL & \bf  NE$\downarrow$ & \bf  SR$\uparrow$ & \bf SPL$\uparrow$  &  \bf  NE$\downarrow$ & \bf  SR$\uparrow$ & \bf CLS$\uparrow$ & \bf nDTW$\uparrow$ & \bf sDTW$\uparrow$  \\
           \midrule
MPM w/ R2R Paths  & 11.99 & 3.44 &  \bf 68.37 &\bf  62.59 & 5.74 & 45.39 & 61.42 & 54.42 & 33.92\\
MPM w/ R4R Paths  & 11.97 & \bf 3.38 & 68.20 & 62.30 &\bf  5.65 & \bf 46.88   & \bf 62.76 & \bf  55.23 & \bf 35.50 \\ 
\bottomrule
\end{tabular}
\caption{MPM performs the best when its collected paths resemble the paths of test environments. Here we only control the lengths of the paths to be similar to the paths of either R2R or R4R.}
\label{tab:path_design}
\end{table*}

\begin{table*}[t]
\small
\centering
  \begin{tabular}{@{}lccccccccccccccc@{}}
    \toprule
     \multirow{2}*{\bf Model} &  \multicolumn{2}{c@{}}{\bf MPM} &      & \multicolumn{4}{c@{}}{\bf R2R Validation Unseen}            & \multicolumn{5}{c@{}}{\bf R4R Validation Unseen}                             \\ 
           &   \bf PT & \bf FT  &\bf TL & \bf  NE$\downarrow$ & \bf  SR$\uparrow$ & \bf SPL$\uparrow$  &  \bf  NE$\downarrow$ & \bf  SR$\uparrow$ & \bf CLS$\uparrow$ & \bf nDTW$\uparrow$ & \bf sDTW$\uparrow$  \\
           \midrule
HAMT+ & \ding{55} & \ding{55} & 11.93 & \bf 3.34 & 67.05 & 61.69 & 5.90 & 44.75  & 61.84 & 54.18 & 33.89  \\
HAMT+ &  \ding{55} &  \ding{51} & 11.84 & 3.40 & 67.65 & 61.74 & 5.83 & 46.35 & \bf 63.66 & \bf 56.73 & \bf 35.87\\
HAMT+   & \ding{51}  &   \ding{51}& 11.99 & 3.44 & \bf 68.37 & \bf 62.59 & \bf 5.65 & \bf 46.88   & 62.76 & 55.23 & 35.50 \\
\bottomrule
\end{tabular}
\caption{Including MPM during both pretraining (PT) and finetuning (FT) can generally lead to the best performance.}
\label{tab:nopt}
\end{table*}

\section{Experiments}
In this section, we present our experimental results with the proposed masked path modeling objective.
\subsection{Settings}
We go through the experimental settings in this part, including our used datasets, evaluation metrics, and implementation details.
\subsubsection{Datasets}
We evaluate the models on different types of VLN datasets, including the Room-to-Room (R2R)~\cite{r2r}, Room-for-Room (R4R)~\cite{jain2019stay} and Room-across-Room (RxR)~\cite{ku2020room} datasets. 

\paragraph{R2R.} The R2R dataset is built based on Matterport3D~\cite{Matterport3D} and has 7,189 paths, with each path paired with 3 different English instructions and the average length of all the paths is 29. R2R is split into training, validation, and test sets; the validation set consists of two splits: 1) \textit{val-seen}, where all the paths are sampled from environments that are also seen in the training set, and 2) \textit{val-unseen}, where paths are sampled from environments that do not appear in the training set so as to test the generalization ability of agents. The paths in the test set are from new environments unseen in the training and validation sets. 

\paragraph{R4R.} The R4R dataset is an algorithmically produced extension of R2R that concatenates two adjacent tail-to-head paths in R2R as well as their corresponding instructions to form a new instruction-path pair. With this extension, R4R has longer paths and instructions, and the paths are not always the shorted path from the starting point to the goal, making the dataset less biased than R2R.

\paragraph{RxR.} The RxR dataset follows the same environment division as that in the R2R dataset. Different from R2R, RxR is a larger dataset that has 16,522 paths in total. In addition, the instructions are multilingual and in three languages, including English, Hindi, and Telugu. The lengths of the instructions in RxR are also much larger than that in R2R (average length: 78 vs. 29).

\subsubsection{Evaluation Metrics} 
We adopt the standard evaluation metrics in VLN~\cite{anderson2018evaluation,jain2019stay,ku2020room} to evaluate models. Specifically, we evaluate models with 1) trajectory lengths (TL): the length of the agent path measured in meters; 2) navigation error (NE): the average distance between the final position of agents and the goal position measured in meters; 3) success rate (SR): the proportion of agents whose final position is within three meters of the target; 4) success rate weighted by normalized inverse path length (SPL): success rate normalized by the ratio between the length of the shortest path and the predicted path.

Because the above metrics are heavily biased towards whether the agent can reach the goal position or not while ignoring the specific path the agents take,~\citet{jain2019stay} propose the Coverage weighted by Length Score (CLS) metric that measures the path fidelity between the predicted path and target path for the R4R dataset. Similarly, \citet{ku2020room} propose normalized dynamic time warping (nDTW) and success rate weighted by dynamic time warping (sDTW)~\cite{Magalhes2019EffectiveAG} for RxR.

\subsubsection{Implementation Details}
\paragraph{Model Architecture.} We build our models upon the HAMT model~\cite{chen2021history} and follow all of its parameter settings except that we use CLIP-ViT~\cite{radford2021learning} pretrained vision encoder and it is not finetuned during training. Specifically, our model consists of a 9-layer text transformer, a 2-layer panoramic transformer for encoding history information, and a 4-layer transformer for cross-modal encoding. In each panoramic observation, there are $K=36$ views of images and we use CLIP-ViT-L-336/14 to encode the input images.

\paragraph{Pretraining.} During pretraining,  we randomly select proxy tasks including masked path modeling for each mini-batch with a predefined ratio as in~\citet{chen2021history}. Different from~\citet{chen2021history}, the CLIP-ViT is frozen instead of finetuned during both pretraining and finetuning in order to save computational costs. We train the model for 200k iterations with the AdamW optimizer~\cite{loshchilov2018decoupled} and the learning rate is set to 5e-5 and the batch size is set to 64. It take around 1 day to finish training on 4 NVIDIA Tesla V100 GPUs. 

\paragraph{Finetuning.} During finetuning, the model is jointly finetuned with the IL+RL and masked path modeling objectives with equal weights. The model is fine-tuned for 300k iterations with a learning rate of 1e-5 and batch size of 8 on a single V100 GPU, taking around 2.5 days to finish.\footnote{Following the hyper-parameters in \url{https://github.com/cshizhe/VLN-HAMT/blob/main/finetune_src/scripts/run_r2r.sh}.} The best model is selected according to performance on the val unseen split. We use the same augmented data as~\citet{hong2021vln} following previous work for the R2R dataset, while no augmented data is used for other datasets. Greedy search is applied in inference following the single-run setting. In both pretraining and finetuning, we do not allow the agents to explore test environments.

\subsection{Main Results}
The main results of the baselines and our model are listed in Table~\ref{tab:r2r},~\ref{tab:r4r}, and~\ref{tab:rxr}. We report both the numbers in the HAMT paper~\cite{chen2021history} and our reproduced performance.

First, it should be noted that because we use the strong CLIP vision encoder, our reproduced HAMT+ baseline can achieve better performance than the original HAMT paper across settings. Especially, on the RxR datasets, our HAMT+ outperforms HAMT by 4.53\% and 4.57 sDTW on the validation seen split and 5.5\% success rate and 4.42 sDTW on the validation unseen split. 

Built upon a strong baseline, our model can still outperform it across settings. Notably, the performance gains are pronounced when measured with path fidelity metrics (i.e., CLS, nDTW, sDTW) on the long-horizon VLN datasets R4R and RxR in the seen environments, indicating that the masked path modeling objective can encourage the models to faithfully follow the natural language instructions and complete the paths accordingly. This is intuitive as the VLN training objectives can optimize the models towards taking the shortest path to reach the final goal, whereas during masked path modeling, the model is trained to reconstruct the original paths, thus the model can follow the instructions more faithfully.

In unseen environments, our model achieves 1.32\%, 2.13\%, and 1.51\% improvements over the baseline in success rates on the R2R, R4R, and RxR datasets respectively on the validation set, demonstrating the effectiveness of our approach. We attribute these improvements to the fact that our objective allows the model to be trained on a variety of paths and can thus improve the generalization ability of the model in unseen environments.

\subsection{Analysis}
In this part, we perform several analyses to gain insights into our proposed method. 
\paragraph{Exploring Unseen Environments.} Because we allow the agents to autonomously acquire data and learn without the need for annotated data, we can train the agents in a scalable way. We hypothesize that when trained with masked path modeling, the agent can be familiarized with the explored environments and thus improve its navigation performance, even though it is not explicitly trained with the VLN objective. To verify this, we train the model with masked path modeling on unseen environments in the validation sets with the same hyper-parameters as before and test its VLN performance. As shown in Table~\ref{tab:explore}, performing masked path modeling on unseen environments can significantly improve the model navigation performance, %\vp{since you use dev data for training now, how do you choose your hyper-parameters?}\zd{same hyper-parameters as before. added this in the text.}
demonstrating the potential of using the objective in a large-scale setting. Especially, exploring unseen environments can bring 4.46\% and 3.56 improvements in SR and sDTW on the R4R validation unseen set respectively.

\begin{figure}[ht]
 \centering
 \includegraphics[width=0.47\textwidth]{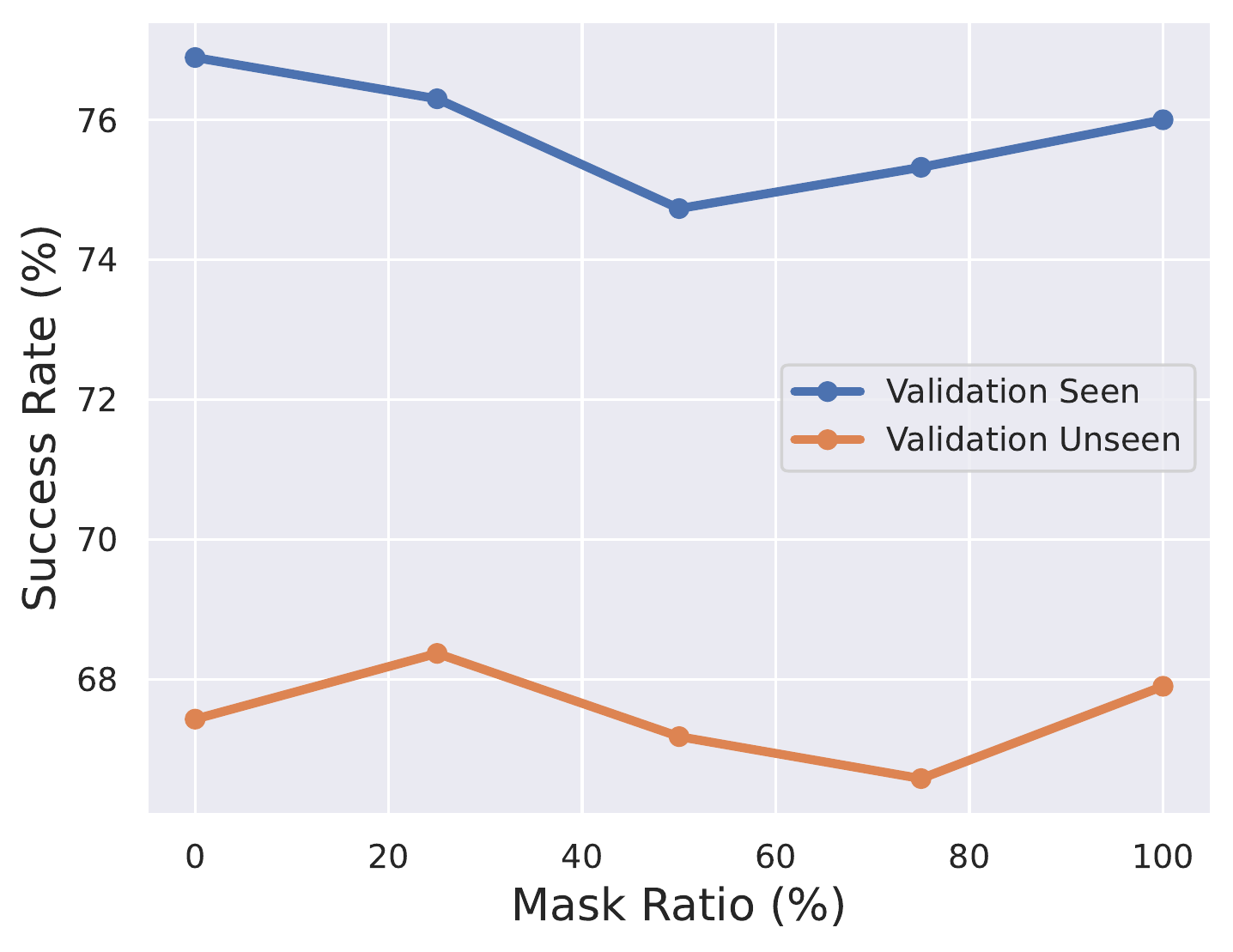}
 \caption{Effect of the mask ratio of the MPM objective. The model is generally robust to this hyper-parameter, with 25\% achieving the best performance on unseen environments. }
 \label{tab:mask}
\end{figure}

\paragraph{Mask Ratio.} The mask ratio is a hyperparameter in masked path modeling and we investigate the optimal ratio in this paragraph. As shown in Figure~\ref{tab:mask}, the objective is generally robust to the mask ratio, with 25\% leading to the best performance and <=50\% bringing improvements on the baseline. Therefore, we choose to randomly mask 25\% of the viewpoints along a path in this paper. 

\paragraph{Path Design.} When collecting the paths, we make the lengths of the sampled paths follow the distribution of that in the training data so that the paths are similar to the navigation paths. As shown in Table~\ref{tab:path_design}, if we switch the path designs between R2R and R4R, the performance gains will drop marginally, indicating that making the paths between MPM and VLN similar can best utilize the MPM objective for VLN. We leave more sophisticated path designs as future work. 

\paragraph{MPM during Pretraining.} During the pretraining stage, we follow HAMT to train the models with masked language modeling, instruction trajectory matching, single-step action prediction and regression, and spatial relationship prediction tasks. We also choose to include the masked path modeling objective during pretraining so as to mitigate the difference between pretraining and finetuning. As shown in Table~\ref{tab:nopt}, we can see that including masked path modeling is important as it can well prepare the models for the finetuning stage, although only doing masked path modeling during finetuning can also bring marginal improvements. Notably, not including MPM during pretraining seems to achieve comparable or even better performance than including it on R4R. One possible explanation is that during pretraining the path lengths are similar to those of R2R, thus the pretrained agent may be more suitable for R2R than R4R.

%\paragraph{Quanlitative Examples.}

\section{Related Work}
In this section, we overview three lines of research, including vision-and-language navigation in general, vision-and-language pretraining with a focus on its applications in VLN, as well as pretraining for control and embodied learning.
\paragraph{Vision-and-Language Navigation.} Building vision-and-language navigation models has received increasing attention in recent years~\cite{anderson2018vision,fried2018unified,wang2018look,li2019robust,zhu2020babywalk,kurita2020generative} and various benchmarks have been proposed to evaluate the ability of embodied agents to follow instructions and accomplish specified tasks~\cite{kolve2017ai2,anderson2018evaluation,savva2019habitat,r2r,chen2019touchdown,ku2020room,shridhar2020alfred,padmakumar2021teach}. In this line of research, representative works include~\citet{fried2018speaker} who propose panoramic action space and use a speaker follow to synthesize additional training data. In addition,~\citet{tan2019learning} propose to mix imitation learning and A3C~\cite{mnih2016asynchronous} and increase the diversity of the synthesized data by adding noise into the environments during data generation. To utilize additional training signals,~\citet{ma2019self} propose the self-monitoring agent that improves vision-language alignment with a co-grounding module and progress monitor;~\citet{zhu2020vision} propose four self-supervised auxiliary tasks that are beneficial for the task of VLN. There are also works on designing better architectures~\cite{chen2021history,chen2022think} and data augmentation strategies~\cite{wang2022less,kamath2023new}.

\paragraph{Vision-and-Language Pretraining.} Pretraining models on large web corpora have proven to be highly effective in natural language processing~\cite{elmo,kenton2019bert,liu2019roberta,brown2020language}, and similar techniques have been applied in computer vision~\cite{chen2020simple,chen2021empirical,he2022masked,bao2022beit} and vision-language communities~\cite{li2019visualbert,chen2020uniter,radford2021learning,kim2021vilt,li2021align,dou2021empirical}. In the field of VLN, researchers have tried to use unimodally or multimodally pretrained language or vision representations~\cite{anderson2018vision,li2019robust}. Notably, the CLIP vision encoder~\cite{radford2021learning} pretrained on large image-caption data has proven to be generally effective for vision-and-language embodied tasks~\cite{khandelwal2022simple,shen2022much}. To jointly learn transferrable vision and language representations for VLN,~\citet{majumdar2020improving} and~\citet{guhur2021airbert} propose to first pretrain models on large image-caption data such as Conceptual Captions~\cite{sharma2018conceptual} and then adapt the representations for VLN tasks by finetuning the model on in-domain VLN data. While the pretrained representations can be useful, the pretraining process does not explicitly connect the learned representations to output actions. To further integrate action generation into VLN pretraining, researchers have attempted to directly use VLN data for pretraining. For example, PREVALENT~\cite{hao2020towards} is pretrained on image-text-action triplets with a single-step action prediction objective and masked instruction modeling objective;~\citet{chen2021history} further propose a single-step regression and spatial relationship prediction objective that introduces more supervisions. However, the pretraining data is limited by the size of VLN data and thus it is difficult to apply their approaches in large-scale settings.

\paragraph{Pretraining for Control and Embodied Learning.} The general idea of pretraining on large-scale data has been adopted in embodied tasks, including data collection~\cite{burda2019exploration}, representation learning~\cite{yang2021representation}, and world model learning~\cite{seo2023masked}. For example,~\citet{pathak2017curiosity} present a curiosity-driven self-supervised data collection approach that encourages agents to explore unfamiliarized states;~\citet{nair2022r3m} pretrain representations on human ego-centric video data and adapt the representations on robotic tasks. Similar to our work,~\citet{wu2023mtm} propose to pretrain models by reconstructing a full trajectory or given a random subset of the same trajectory, and the pretrained models can be used for different downstream
purposes like inverse dynamics, forward dynamics, imitation learning, offline RL, and representation learning. Different from these works, our method explicitly optimizes models for conditioned action generation and the agent can self-collect a rich amount of data for pretraining.

\section{Conclusion}
In this paper, we propose masked path modeling, a pretraining objective designed for vision-and-language navigation. The objective can utilize scalable data explored from different environments as well as improve the agent's conditioned action generation ability. We incorporate the objective into a strong baseline and demonstrate improvements across different settings. An analysis also reveals the potential for scaling the objective to large-scale settings. Future directions include designing better exploration strategies as well as investigating applications in more fields.

\bibliography{anthology,custom}
\bibliographystyle{acl_natbib}

%\appendix

%\section{Example Appendix}
%\label{sec:appendix}

%This is a section in the appendix.

\end{document}